\documentclass[conference]{IEEEtran}
\IEEEoverridecommandlockouts
\usepackage{cite}
\usepackage{amsmath,amssymb,amsfonts}
\usepackage{algorithmic}
\usepackage{graphicx}
\usepackage{textcomp}
\usepackage{xcolor}
\usepackage{ctable}
\usepackage{caption}
\usepackage{subcaption}
\usepackage{makecell}
\usepackage{multirow} 
\usepackage{xurl}
\usepackage[super]{nth}
\usepackage[version=4]{mhchem} 
\usepackage{enumitem} 

\makeatletter
\newcommand{\linebreakand}{%
  \end{@IEEEauthorhalign}
  \hfill\mbox{}\par
  \mbox{}\hfill\begin{@IEEEauthorhalign}
}
\makeatother

\usepackage{soul}
\soulregister\cite7

\usepackage[margin=1.5cm]{geometry}
\usepackage{lipsum}

\def\BibTeX{{\rm B\kern-.05em{\sc i\kern-.025em b}\kern-.08em
    T\kern-.1667em\lower.7ex\hbox{E}\kern-.125emX}}

\date{June 2025}

\begin{document}
\title{Generalizing Scaling Laws for Dense and Sparse Large Language Models}

\author{\IEEEauthorblockN{Md Arafat Hossain}
\IEEEauthorblockA{\textit{Department of Computer Science} \\
\textit{Iowa State University} \\
Email: arafat@iastate.edu 
 }
\and
\IEEEauthorblockN{Xingfu Wu, Valerie Taylor}
\IEEEauthorblockA{\textit{Mathematics and Computer Science Division} \\  
\textit{Argonne National Laboratory} \\
Email: \{xingfu.wu, vtaylor\}@anl.gov
 }
\and
\IEEEauthorblockN{Ali Jannesari}
\IEEEauthorblockA{\textit{Department of Computer Science} \\
\textit{Iowa State University} \\
Email: jannesar@iastate.edu 
 }
}
\maketitle

\begin{abstract}
Despite recent advancements of large language models (LLMs), optimally predicting the model size for LLM pretraining or allocating optimal resources still remains  a challenge. Several efforts have addressed the challenge by proposing different empirical scaling laws, but almost all of them are architecture-specific (dense or sparse). In this work we revisit existing empirical scaling laws and propose a generalized scaling law to provide a unified framework that is applicable to both dense and sparse large language models. We evaluate and compare our proposed scaling law with existing scaling laws and demonstrate that our proposed scaling law captures the scaling behavior of existing scaling laws. Further, we show an IsoFLOP comparison between our proposed scaling law and the state-of-the-art scaling law to illustrate the effectiveness of our proposed scaling law for Mixture-of-Expert (MoE)-based very large LLMs like DeepSeek-V3. Our proposed scaling law can be used to estimate the best model hyperparameters (Model size, Tokens and Compute) for a given sparsity or to identify the optimal sparsity for the given model hyperparameters.


\end{abstract}


\section{Introduction} \label{intro}

In recent years, transformer architectures \cite{vaswani2017attention} have revolutionized the deep learning approach especially the field of natural language processing with the development of transformer-based architectures \cite{vaswani2017attention} 
such as BERT \cite{devlin2019bert} and GPT-3 \cite{brown2020language} etc. Transformer architecture is now the foundation for the majority of popular large language models (LLMs). Since the release of ChatGPT LLMs have gained more attention across diverse domains as a key innovative technology.  With the growing popularity across diverse domain there is an increasing need to scale LLM knowledge and accuracy while also reducing the inference latency and training cost. However, achieving these goals is fundamentally challenging due to the computational structure of standard transformer models, as computation grows rapidly with the increase of model size and context length. 

In traditional transformer architecture, parameters in every layer are activated to process every token. Each token passes through the same feed-forward network, and in a multi-head attention setting, every attention considers every other position. These traditional forms of transformers are called dense transformers. The complex and exhaustive architecture allows the dense models to understand the rich contextual meaning of the input dataset, but it comes with a high computational cost that scales with both model size and sequence length.

One way to reduce the cost is to introduce sparsity where all the available parameters are not used to process a token in a sparse model. There are several ways to introduce sparsity, but in this work we focus on two methods:  pruning~\cite{cheng2023survey} and mixture of experts (MoE) ~\cite{shazeer2017outrageously}. Pruning-based sparsity permanently removes parameters from the model. Pruning can be categorized into three main groups \cite{cheng2023survey} namely: structured pruning, unstructured pruning, and semi-structured pruning. In  structured pruning,  entire higher-level units or even whole layers are removed; this process delivers universal speedup but sometimes at the cost of accuracy.  In unstructured pruning, individual weights are removed from anywhere in the network; usually the least important weights are pruned. 
In semi-structured pruning, one tries to strike a balance between unstructured and structured pruning approaches by enforcing fixed pruning patterns in small regions of the network. On the other hand,  in MoE-based models the feed-forward network for a transformer model is replicated multiple times, and each of these copies is called an ``expert". In the MoE-based models only a fraction of experts are activated per token, instead of all the experts and that's how sparsity is introduced. The number of activated experts is a fixed value and determines the sparsity level. The tokens are routed to different experts during the pretraining process. The size of the experts is usually equal to the size of the feed-forward network but can differ too. 

Several models \cite{rae2021scaling,smith2022using, thoppilan2022lamda, chowdhery2023palm, olmo20242} have been developed over the past few years with traditional dense architecture. But lately, sparse transformers\cite{child2019generating}---have become extremely popular, and several models \cite{du2022glam, fedus2022switch,zoph2022designing, muennighoff2024olmoe,liu2024deepseek, dai2024deepseekmoe, jiang2024mixtral, guo2025deepseek} have been developed using the concept of sparsification.  One common factor among all these models is that as they are evolving, their size has been increasing significantly from GPT-2 with $1.5$ billion parameters \cite{radford2019language}  to PaLM with 540 billion parameters~\cite{chowdhery2023palm}, to DeepSeek-V3, a strong mixture-of-experts (MoE) language model with 671 billion parameters, 37 billion of which are activated for each token \cite{deepseekai2025deepseekv3technicalreport}, which has 94.49\% sparsity. The reason is, it has been proven that as model size, dataset size, and the amount of compute used for training increase, LLM performance improves smoothly \cite{kaplan2020scaling}. However with the increase of size, these models require large training datasets, eventually increasing the compute budget \cite{kaplan2020scaling, hoffmann2022training} to  several petaFLOP/s-days of computation to achieve optimal performance. This makes LLM training time-consuming and resource intensive. As a result, determining the optimal allocation of resources, model size, and training data and estimating the training performance before conducting the actual training is crucial.
 
 Scaling laws are the empirical relationships, often expressed as power laws---describe a model's performance based on allocated resources. These laws help determine the optimal model size and data volume for a given  compute budget to achieve optimal pretraining performance. Several empirical scaling laws based on different LLM architectures have been proposed in recent years \cite{kaplan2020scaling, hoffmann2022training, frantar2023scaling, abnar2025parameters, henighan2020scaling, clark2022unified, ludziejewski2024scaling,  gadre2024language}. Few scaling law works focused on training precision  \cite{kumar2024scaling}, distillation \cite{busbridge2025distillationscalinglaws}, and quantization \cite{chen2025scaling}. Several studies have also been done on  vision \cite{frantar2023scaling}, audio \cite{droppo2021scaling,qiu2024efficient}, and mixed-modal language models \cite{aghajanyan2023scaling}. This  interest in scaling laws for different domains makes them pivotal in deep neural network research. However, until now, nearly all existing approaches have been architecture-specific (dense or sparse) and also use different scaling law coefficients (details in Section \ref{scaling_laws}) which makes it difficult to perform cross-architecture comparisons and budget planning. In this work we show that transformers, regardless of dense and sparse architectures, exhibit a common scaling behavior which depends on the number of active/nonzero parameters. To demonstrate that, we generalize the existing empirical scaling laws to propose a scaling law that enables accurate performance predictions and optimal resource allocation across architectures. In this paper we focus on the scaling laws for both dense and sparse LLMs. We make the following contributions:

 \begin{itemize}
 \item We discuss the existing empirical scaling laws for dense and sparse LLMs and their limitations (Section II).
\item We propose a generalized scaling law for both dense and sparse LLMs and discuss the relationship between the proposed scaling law and the existing laws (Section III).
\item We evaluate and compare the performance of the proposed scaling law and the existing laws using some test models to demonstrate its effectiveness, and we present hyperparameter optimization results for some existing scaling laws (Section IV). 
\end{itemize}

Section V summarizes this work and suggests avenues for potential future research.  

\section{Scaling Laws for LLMs}\label{scaling_laws}
Based on the number of parameters available in the architecture and the number of parameters that are being used during pretraining, transformer models can be categorized into dense and sparse transformers. Previous works \cite{rae2021scaling,smith2022using, thoppilan2022lamda, chowdhery2023palm, olmo20242, du2022glam, fedus2022switch,zoph2022designing, muennighoff2024olmoe,liu2024deepseek, dai2024deepseekmoe, jiang2024mixtral, guo2025deepseek} have shown that these models exhibit different behavior during pretraining as they scale, and hence their scaling behaviors are studied separately.  

\subsection{Scaling Laws for Dense Models}
 Traditional transformer models are dense in nature. Kaplan et al.~\cite{kaplan2020scaling} studied the behavior of dense LLMs with a series of experiments by monitoring different hyperparameters: model size, model shape, total compute, data volume, and batch size during pretraining. The authors established that model loss depends strongly on scale, which consists of the number of model parameters $N$, the dataset size (number of tokens) $D$, and the amount of compute $C$ used for training. It also depends mildly on other architectural hyperparameters such as model shape: depth and width. For dense models, the compute $C$ (training FLOPs) required to train a transformer model with $N$ parameters and $D$ tokens is $C=6ND$ \cite{kaplan2020scaling} because approximately 6 floating point operations are needed for every parameter in the model and every training
sample. The authors also suggested that if compute is increased by $10$ times, then model size should be increased by  $5.5$ times and dataset size  by $1.8$ times to avoid overfitting. They represented training loss as a function of number of parameters and data size and proposed the empirical scaling law:

 \begin{equation}
     L(N, D) = \left[ \left(\frac{N_C}{N} \right)^{\frac{\alpha_N}{\alpha_D}} + \frac{D_C}{D}\right ]^{\alpha_D}
     \label{kaplan}
 \end{equation}
 
 The best-fitted values for the coefficients and exponents in Equation \ref{kaplan} are listed in Table \ref{tab:kaplan_tab}. 
 
\begin{table}[ht]
    \centering
    \begin{tabular}{c c c c c}
    \specialrule{.1em}{0em}{0em}
    Coefficients & $\alpha_N$ & $\alpha_D$ & $N_C$ & $D_C$ \\
    \hline
    Values & 0.076 & 0.103 & $6.4\times 10^{13}$ & {$1.8 \times 10^{13}$} \\
    \specialrule{.1em}{0em}{0em}
    \end{tabular}
    \caption{Fitted parameter values for Equation \ref{kaplan}} 
    \label{tab:kaplan_tab}
\end{table}
 
 Using three different approaches---fixed model size with varying numbers of training tokens, IsoFLOP profiling, and parametric function fitting---Hoffmann et al. ~\cite{hoffmann2022training} extended and corrected Equation 1 in \cite{kaplan2020scaling} and concluded that model size and number of tokens should be increased in proportion to avoid overfitting. 
 The authors represented training loss as a combination of irreducible loss or entropy $e$, a parameter-dependent term or the loss incurred by a model of fixed parameter size, and a data size term or the loss induced by a fixed number of training steps. Based on the observations from their three experiments, the authors proposed Equation \ref{hoffman} as their scaling law.
\vspace{-0.5em}
 \begin{equation}
     L(N, D) = e + \frac{a}{N^\alpha}+\frac{b}{D^\beta}
     \label{hoffman}
 \end{equation}

 \begin{figure}
    \centering
    \includegraphics[width=0.6\linewidth]{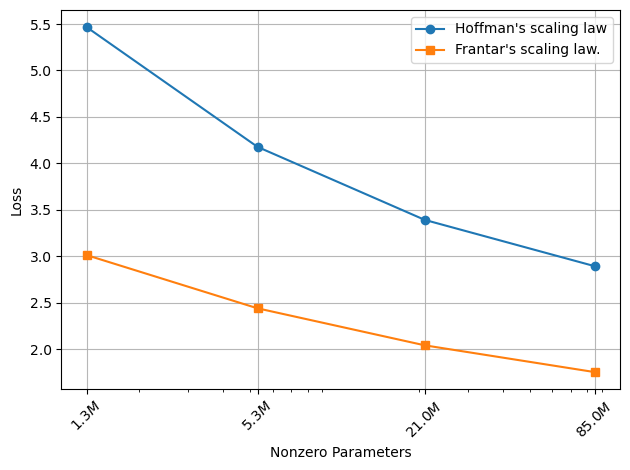}
    \caption{Comparison between Hoffman scaling law and Frantar scaling law at $0\%$ sparsity.}
    \label{fig:frantar_vs_hoffman}
\end{figure}

This law establishes a power-law relationship between loss $L$,  $N$, and $D$. Here $e$ captures the entropy of the natural text, which is the minimum loss as both $N$ and $D$ approach infinity. The second term (parameter-dependent term) captures the inverse relationship between $N$ and loss. The third term (data-dependent term) captures the impact of  $D$ on loss. The constants $a$ and $b$ and the exponents $\alpha$ and $\beta$ are determined empirically through some experiments and fitting the data. Its best-fitted values for the coefficients and exponents are listed in Table \ref{tab:hoffman_tab}. Equation \ref{hoffman} currently is the state-of-the-art empirical scaling law for dense LLMs. For the rest of the paper, we will refer Equation \ref{hoffman} as Hoffman scaling law. 

\begin{table}[h]
    \centering
    \begin{tabular}{c c c c c c}
    \specialrule{.1em}{0em}{0em}
    Coefficients & $e$ & $a$ & $b$ & $\alpha$ & $\beta$\\
    \hline
    Values & $1.69$ & $406.4$ & $410.7$ & $0.34$ & $0.28$\\
    \specialrule{.1em}{0em}{0em}
    \end{tabular}
    \caption{Fitted parameter values for Equation \ref{hoffman}} 
    \label{tab:hoffman_tab}
\end{table}

\subsection{Scaling Laws for Sparse Models} \label{sparse_scaling_laws}
Sparse LLMs use only a subset of the parameters available in the models. 
Several works had proposed scaling laws for sparse transformer models \cite{frantar2023scaling, abnar2025parameters, ludziejewski2024scaling}. These scaling laws are based on different sparsification methods.  
In this work we focus on two types of sparsifications: pruning and mixture-of-experts (MoE).

\begin{figure}[ht]
    \centering
    \includegraphics[width=0.6\linewidth]{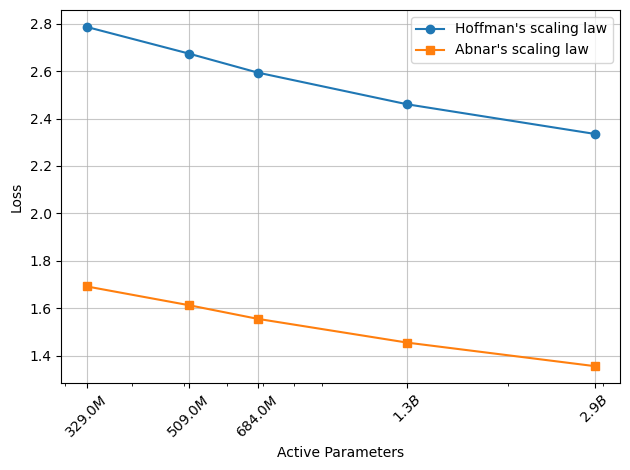}
    \caption{Comparison between Hoffman scaling law and Abnar scaling law at $0\%$ sparsity.}
    \label{fig:hoffman_vs_abnar}
\end{figure}

\subsubsection{Pruned Models} One way to introduce sparsity in transformer models is pruning, where a subset of parameters are removed or masked and  are not used during the forward pass or backpropagation, hence making no contribution to the training process.  
Frantar et al.~\cite{frantar2023scaling} studied the scaling behavior of models pruned by unstructured pruning. In their work, the authors defined sparsity, 
$S = \frac{\text{Total Parameters} - \text{Nonzero Parameters}}{\text{Total Parameters}}$.
Based on the pretraining results of $48$ models with different sparsity levels, numbers of nonzero parameters and training tokens, the authors claimed that sparsity introduced by pruning affects only the parameter-dependent term of Hoffman scaling law and has minor to no impact on the data-dependent term. The authors also claim that increasing sparsity leads to a better pretraining loss. Their experimental results also showed that as the number of nonzero parameters and training tokens—in other words, total compute increased the pretraining loss keeps getting better, which holds the claim of Hoffman scaling law \cite{hoffmann2022training}. Using the Hoffman scaling law as the foundation, the authors fitted loss as a function of nonzero parameters $N$, data $D$, and sparsity $S$ as shown in Equation \ref{frantar}.
\begin{equation}
    L(N, D, S) = \left( a_S (1-S)^{b_S} +c_S \right).\left(\frac{1}{N}\right)^{b_N} + \left(\frac{a_D}{D} \right)^{b_D} + c
    \label{frantar}
\end{equation}

Following the similar approach of \cite{hoffmann2022training} the authors presented the best-fitted values for the scaling law coefficients of Equation \ref{frantar}. Table \ref{tab:frantar_tab} lists the scaling law parameter values mentioned in  \cite{frantar2023scaling}. For the rest of the paper, we will refer to Equation \ref{frantar} as the Frantar scaling law. 

\begin{table}[ht]
    \centering
    \resizebox{0.49\textwidth}{!}{
        \begin{tabular}{c c c c c c c c}
        \specialrule{.1em}{0em}{0em}
        Coefficients & $a_S$ & $b_S$ & $c_S$ & $b_N$ & $a_D$ & $b_D$ & $c$\\
        \hline
        Values & $16.8$ & $0.722$ & $45$ & $0.245$ & $6.90\times 10^8$ & $0.203$ & $0.651$\\
        \specialrule{.1em}{0em}{0em}
        \end{tabular}
    }
    \caption{Fitted parameter values for Equation \ref{frantar}}
    \label{tab:frantar_tab}
\end{table}

To draw the similarities between Hoffman scaling law in Equation \ref{hoffman} and Frantar scaling law in Equation \ref{frantar}, we used the similar symbols used in Hoffman scaling law to reformat the scaling law coefficients in Frantar scaling law. We used $e$ instead of $c$ to capture the lower bound on the loss that represents the inherent stochasticity of the modeling problem. We replaced $b_N$ and $b_D$ with $\alpha$ and $\beta$, respectively, and ${a_D}^{b_D}$ with $b$. $c_S$ is the constant sparsity factor. The reformatted version of Frantar scaling law is Equation \ref{frantar_modified}.
%
\begin{equation}
    L(N, D, S) = e + \left( a_S (1-S)^{b_S} +c_S \right) \frac{1}{N^\alpha} + \frac{b}{D^\beta}
    \label{frantar_modified}
\end{equation}

Equation \ref{frantar_modified} helps visualize the transition from the dense Hoffman scaling law Equation \ref{hoffman} to the sparse Frantar scaling law Equation \ref{frantar} in order to investigate which factor is directly impacted by sparsity. Based on the changes in representation, we recalculate the best-fitted values for the scaling law coefficients in the reformatted Frantar scaling law, and the values are listed in Table \ref{tab:frantar_tab_mod}.

\begin{table}[ht]
    \centering
    \begin{tabular}{c c c c c c c c}
    \specialrule{.1em}{0em}{0em}
    Coefficients & $a_S$ & $b_S$ & $c_S$ & $b$ & $\alpha$ & $\beta$ & $e$\\
    \hline
    Values & $16.8$ & $0.722$ & $45$ & $62.271$ & $0.245$ & $0.203$ & $0.651$\\
    \specialrule{.1em}{0em}{0em}
    \end{tabular}
    \caption{Fitted parameter values for Equation \ref{frantar_modified}}
    \label{tab:frantar_tab_mod}
\end{table}

While Equation \ref{frantar_modified}  functionally captures the effect of sparsity on loss, the maximum sparsity used in the experiments was $87.5\%$~\cite{frantar2023scaling} while other works \cite{abnar2025parameters} show that better performance gain can be achieved by introducing even more sparsity (even at 98\% sparsity). 
The largest model used in Frantar scaling law experiments had only $85M$ nonzero parameters which are relatively much smaller. Although this work used the Hoffman scaling law as baseline, at $0\%$ sparsity or for dense models with the same dataset, Equation \ref{frantar_modified}  provides different loss predictions than those in Hoffman scaling law with an average MSE (Mean Squared Error) of $3.04$, as shown in Figure \ref{fig:frantar_vs_hoffman}. Thus, we can conclude that Hoffman scaling law is not a special case of Frantar scaling law where sparsity is $0\%$


      


\begin{figure}[ht]
  \centering
    \includegraphics[width=0.6\linewidth]{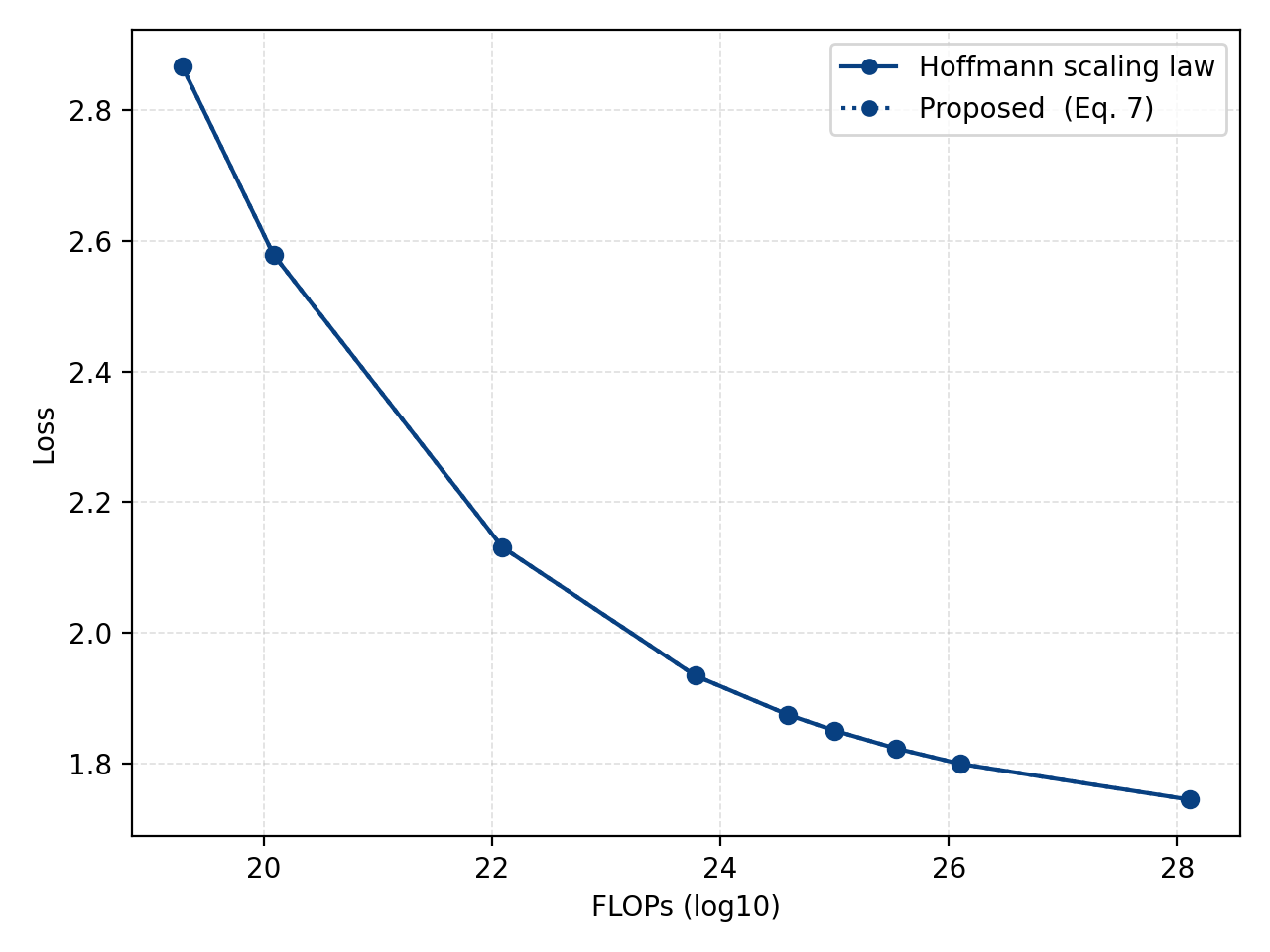}

  \caption{Loss prediction of Hoffman scaling law and the proposed scaling law.}
  \label{fig:hoffman_vs_proposed_comparison}
\end{figure}

\subsubsection{MoE Models} Another way to introduce sparsity is to use the mixture-of-experts (MoE) method \cite{shazeer2017outrageously} where the feed-forward network of a transformer model is replicated multiple times and each of these copies is called an ``expert." Only a fraction of experts are activated per token. 
The tokens are routed to different experts during the pretraining process. 

 \begin{figure*}[ht]
    \begin{subfigure}[t]{0.33\textwidth}
    \centering
    \includegraphics[width=\linewidth]{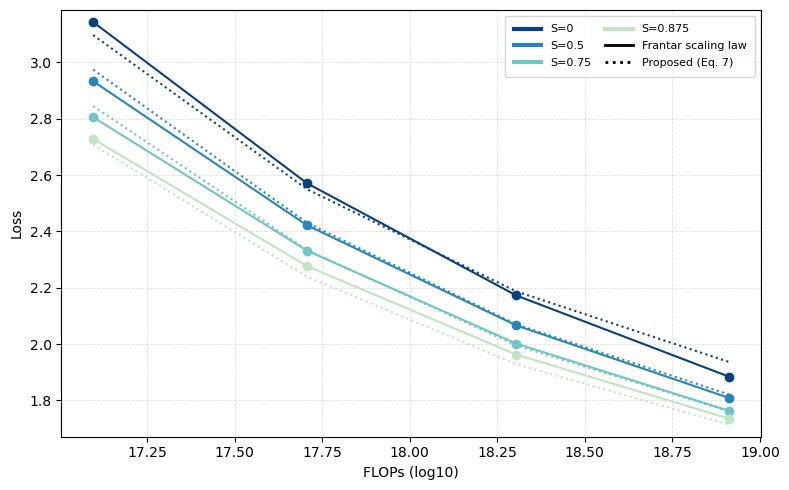}
    \subcaption{Training tokens: $16B$}
  \end{subfigure}\hfill
  \begin{subfigure}[t]{0.33\textwidth}
    \centering
    \includegraphics[width=\linewidth]{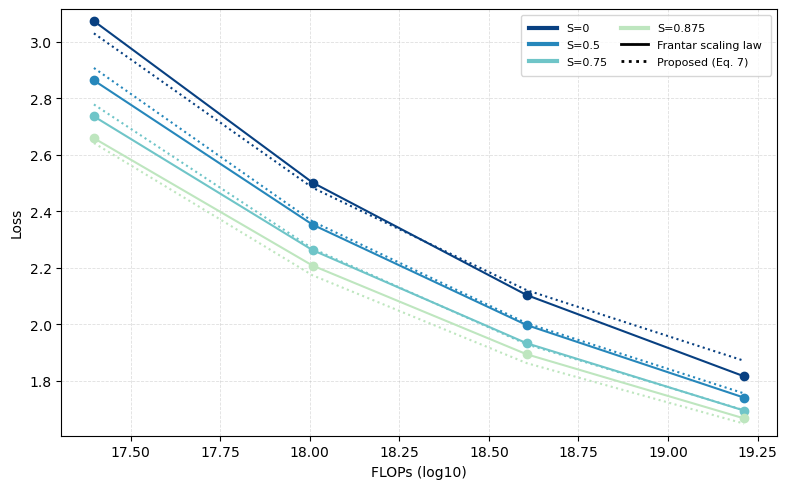}
    \subcaption{Training tokens: $32B$}
  \end{subfigure}
    \begin{subfigure}[t]{0.33\textwidth}
    \centering
    \includegraphics[width=\linewidth]{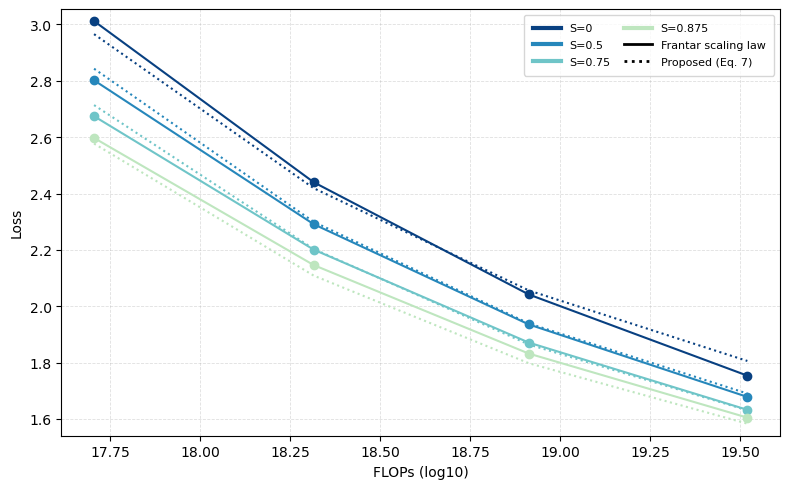}
    \subcaption{Training tokens: $65B$}
  \end{subfigure}
  \caption{As sparsity and the number of nonzero/active parameters increase, the pretraining loss decreases. In each figure, the number of non-zero parameters in the models used ranged from $1.3M-85M$ with varying sparsity levels, and the number of training tokens is mentioned in the subcaption.}
  \label{fig:frantar_vs_proposed_comparison}
\end{figure*} 

Abnar et al.~\cite{abnar2025parameters} investigated the sparse MoEs through a large-scale empirical study. The authors defined sparsity, $S=\frac{E-K}{E}$, where $E$ is the total number of experts and $K$ is the number of active experts. For five ($3e19, 6e19, 1e20, 3e20, \text{ and } 1e21$) different compute budgets and seven ($0\%,25\%,50\%,75\%,90\%,95\%, \text{ and } 98\%$) different sparsities, different models were trained. The conclusion they made was that as we increase sparsity along with model size and number of training tokens, the pretraining loss decreases. For IsoFLOP experiments, they noticed that each for a specific sparsity level, each model size has a sweet spot. Until that spot, the pretraining loss keeps decreasing, and beyond that spot, pretraining loss keeps increasing. Based on the experimental results, the authors proposed Equation \ref{abnar} as their scaling law. For the rest of the paper, we will refer to Equation \ref{abnar} as Abnar scaling law.

\begin{equation}
    L(N,D,S) = e + \frac{a}{N^\alpha} + \frac{b}{D^\beta} + \frac{c}{(1-S)^\lambda} + \frac{d}{(1-S)^\delta N^\gamma}
    \label{abnar}
\end{equation}

Through a grid initialization, for a sparsity of $98\%$, the best-fitted coefficient values in Equation \ref{abnar} are determined empirically through the experiments and fitting the data shown in Table \ref{tab:abnar_tab}. 

\begin{table}[ht]
    \centering
    \begin{tabular}{c c c c c c c c c c c}
    \specialrule{.1em}{0em}{0em}
    Coefficients & $e$ & $a$ & $b$ & $c$ & $d$ \\
    \hline
    Values & $0.94$ & $16612.50$ & $5455.67$ & $0.4598$ & $17.26$\\
    \hline
    Coefficients & $\alpha$ & $\beta$ & $\lambda$  & $\delta$ & $\gamma$\\
    \hline
    Values & $0.5962$ & $0.3954$ & $-0.1666$ & $0.1603$ & $0.1595$\\
    \specialrule{.1em}{0em}{0em}
    \end{tabular}
    \caption{Fitted parameter values for Equation \ref{abnar}}
    \label{tab:abnar_tab}
\end{table} 

Abnar scaling law introduced two different terms in the scaling law along with the previous three terms from Equation \ref{hoffman} \cite{hoffmann2022training} to capture the impact of sparsity. The newly introduced terms are interdependent. The fifth term $\frac{d}{(1 - S)^\delta N^\gamma}$ in Equation \ref{abnar} can be interpreted as an active parameter term because the denominator term is a multiplication of $N$ and $1-S$. As the total number of parameters already includes the active parameters, having two different terms for total parameters and active parameters in Equation \ref{abnar} can be deemed as redundant. While this may help the Abnar scaling law to fit the experimental results better, we later show that this can be represented in a generalized way without introducing redundancy.  
 
 Abnar scaling law also used Hoffman scaling law as the base. We compared both of these laws at $0\%$ sparsity using the same model size and data size used in Abnar scaling law experiments. The results were off from each other with and average MSE of $1.06$ in Figure \ref{fig:hoffman_vs_abnar}.

\section{Generalizing Scaling Laws}\label{generalization}

We observed that regardless of the dense or sparse architecture, all the previous works mentioned so far use the number of nonzero or active parameters to measure the total compute. Hoffman et al. \cite{hoffmann2022training} used $C=6ND$ where $N$ is the total number of parameters and for dense models this is the number of nonzero/active parameters. Abnar et al. \cite{abnar2025parameters} used $C=6N_aD$ where $N_a$ is the number of active parameters. Frantar et al. \cite{frantar2023scaling} used $C=6ND\cdot c_{mul}(S) $ where the authors used dense Hoffman's compute formula and multiplied it by a sparsity-dependent factor $c_{mul}(S)$.  
We have also observed in the preceding section that the mentioned sparse scaling laws had their limitations and do not replicate the dense scaling law behavior when the sparsity is $0\%$. In Section \ref{intro} we have also discussed that due to the architecture-specific nature of existing scaling laws where some scaling laws take total parameters into account for their scaling law and some use the number of nonzero/active parameters, cross-architecture comparison and budget planning become challenging. On top of that, different scaling laws use different scaling law coefficients that need to be tuned carefully to fit the early experimental results for better prediction. We think that the scaling behavior, regardless of the architecture (dense or sparse), is dependent on the number of active/non-zero parameters and propose a generalized scaling law that can address the earlier-mentioned issues.

This proposed scaling law should  
1) be able to describe the scaling behavior for both dense and sparse models; 2) achieve similar predictions and capture the scaling behavior observed in the experiments of \cite{kaplan2020scaling, hoffmann2022training, frantar2023scaling,abnar2025parameters} as increasing model size, data size, and compute; and 3) capture the behavior for sparse models as observed in \cite{frantar2023scaling,abnar2025parameters} for increasing the sparsity. Furthermore, for $0\%$ sparsity or dense models, the scaling law should behave exactly as the Hoffman scaling law. 

\begin{figure}[ht]
    \centering
    \includegraphics[width=0.8\linewidth]{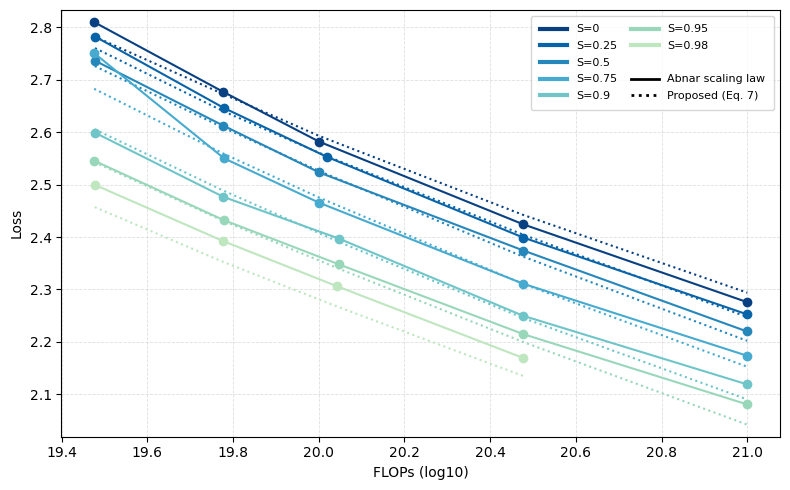}
    \caption{Scaling behavior and prediction comparison between Abnar scaling law and our proposed scaling law.}
    \label{fig:ax_abnar_comparsion}
\end{figure}

For simplicity, we use the term ``active parameters" ($N$) for active/nonzero parameters in the MoE and pruned models. For the dense models, the total number of active parameters is the total number of parameters. We use Hoffman scaling law (Equation \ref{hoffman}) as the base to propose a generalized scaling law. 
We define the sparsity as follows: 
\begin{equation}
    S=\frac{\text{Total Parameters - Active Parameters}}{\text{Total Parameters}} 
\end{equation}

 Where $0 \le S<1$ (The number of active parameters in a model cannot be zero and hence, $S\not = 1$). $\frac{N}{1-S}$ is the total number of parameters. In the parameter-dependent second term of Hoffman scaling law in Equation \ref{hoffman}, we replace $N$ with $\frac{N}{1-S}$; the parameter-dependent term thus becomes $\frac{a}{\left({\frac{N}{1-S}}\right)^\alpha}$, which is simplified as $\frac{a (1-S)^\alpha}{N^\alpha}$. Similar to the Frantar et al. approach in \cite{frantar2023scaling} we introduce an upper-bound constant sparsity factor $c$; but unlike Frantar scaling law, this factor depends on the sparsity  $S$. So we define the parameter-dependent term as $\left(a (1-S)^\alpha + c\cdot S\right)\frac{1}{N^\alpha}$. 

While $e$ corresponds to irreducible loss, empirical fits over $N$ and $D$ treat the constant as an effective offset. For sparse models, this offset can vary systematically with sparsity. We model this dependence with minimal multiplicative factor $(1-S)^\gamma$, preserving the dense limit and power law behavior of scaling laws. Accumulating these changes, we propose Equation \ref{ax} as our proposed scaling law. 

\begin{equation}
    L(N,D,S) = e (1-S)^\gamma+ \left(a (1-S)^\alpha + c\cdot S\right)\frac{1}{N^\alpha} +\frac{b}{D^\beta}
    \label{ax}
\end{equation}

This equation quantifies the impact of sparsity on loss. Notice that we did not change the values of the data-dependent term ($\frac{b}{D^\beta}$) in Equation \ref{hoffman} (Hoffman scaling law), because the work in \cite{frantar2023scaling,abnar2025parameters} has shown experimentally that sparsity has no effect on this term. 



\begin{figure*}[ht]
    \begin{subfigure}[t]{0.49\textwidth}
    \centering
    \includegraphics[width=\linewidth]{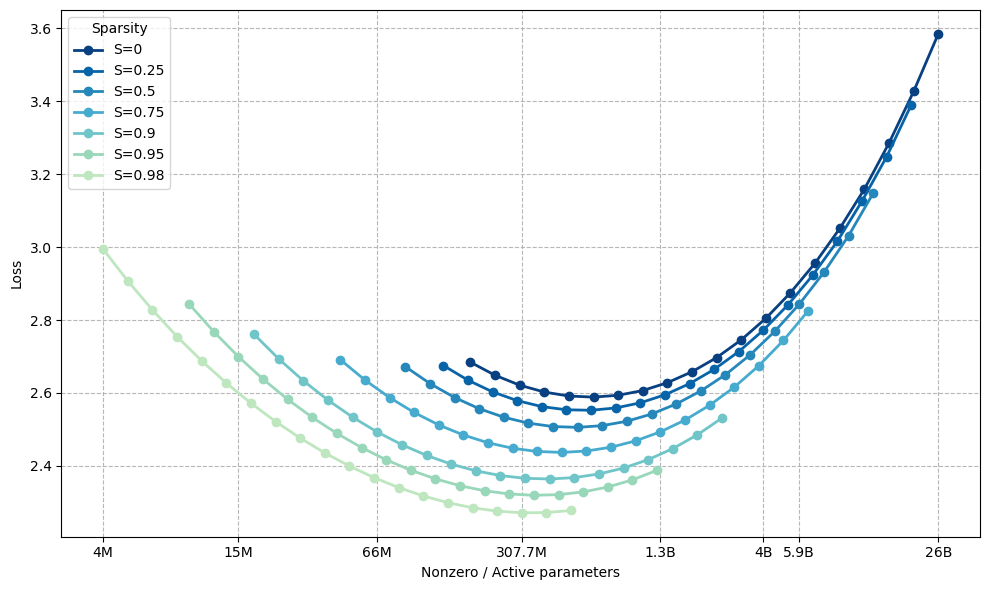}
  \end{subfigure}\hfill
  \begin{subfigure}[t]{0.49\textwidth}
    \centering
    \includegraphics[width=\linewidth]{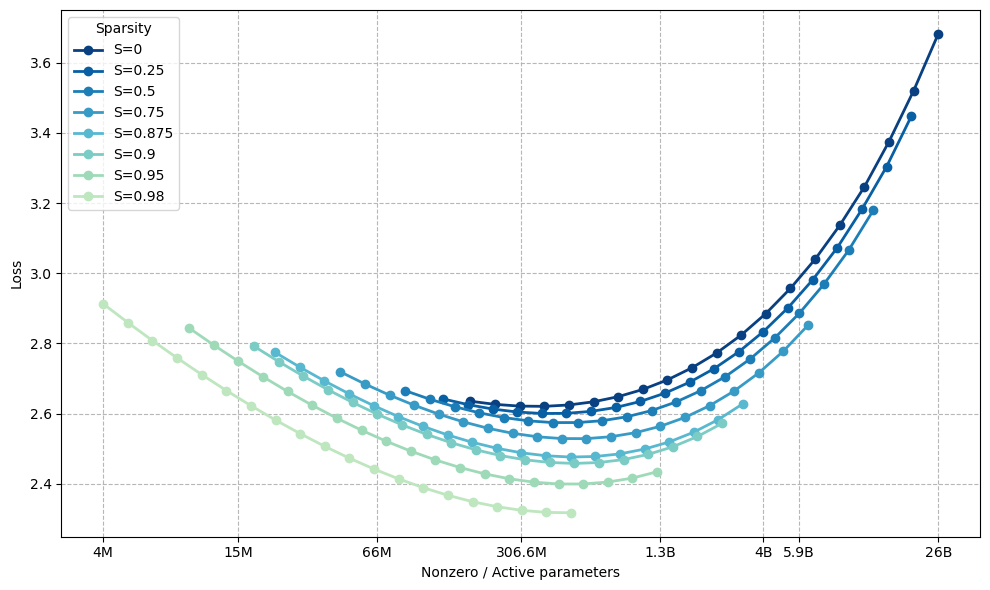}
  \end{subfigure}
  \caption{IsoFLOP plot of Abnar scaling law (left) and our proposed scaling law (right) for a compute budget of $1e20$ FLOPs.}
  \label{fig:isoflop_comparison}
\end{figure*}

\textbf{Lemma}: If sparsity $S$ is 0, Equation \ref{ax} is the Hoffman scaling law in Equation \ref{hoffman}.

If the sparsity $S$ is 0,  the total number of parameters is equal to the total number of active parameters. From Equation \ref{ax} we have
\vspace{-0.5em}
\begin{equation}
    L(N,D,0) = e (1-0)^\gamma+ \left(a (1-0)^\alpha + c\cdot 0\right)\frac{1}{N^\alpha} +\frac{b}{D^\beta} 
    \label{ax0}
\end{equation}
Then, simplifying Equation \ref{ax0}, we have
\begin{equation*}
    L(N,D,0)= e + \frac{a}{N^\alpha} +\frac{b}{D^\beta}. 
\end{equation*}

From Equation \ref{hoffman}, we have 
\vspace{-0.5em}
\begin{equation*}
    L(N,D)= e + \frac{a}{N^\alpha} +\frac{b}{D^\beta}. 
\end{equation*}

Because the coefficients $e, a, b, \alpha, \beta$ are the same, we have 

$L(N,D, 0)-L(N,D) = 0$.

\begin{figure*}[ht]
    \begin{subfigure}[t]{0.49\textwidth}
    \centering
    \includegraphics[width=\linewidth]{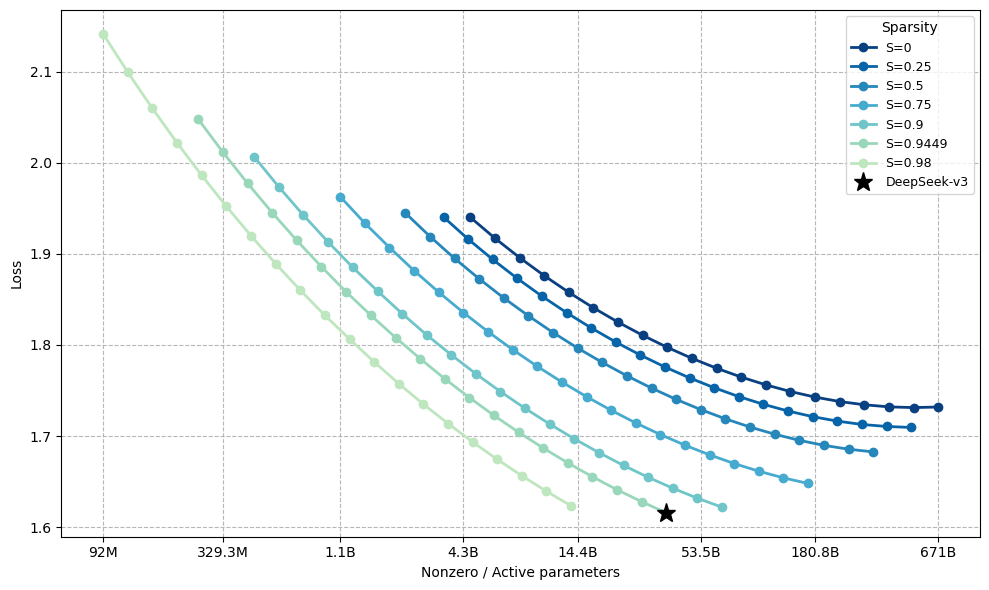}
  \end{subfigure}\hfill
  \begin{subfigure}[t]{0.49\textwidth}
    \centering
    \includegraphics[width=\linewidth]{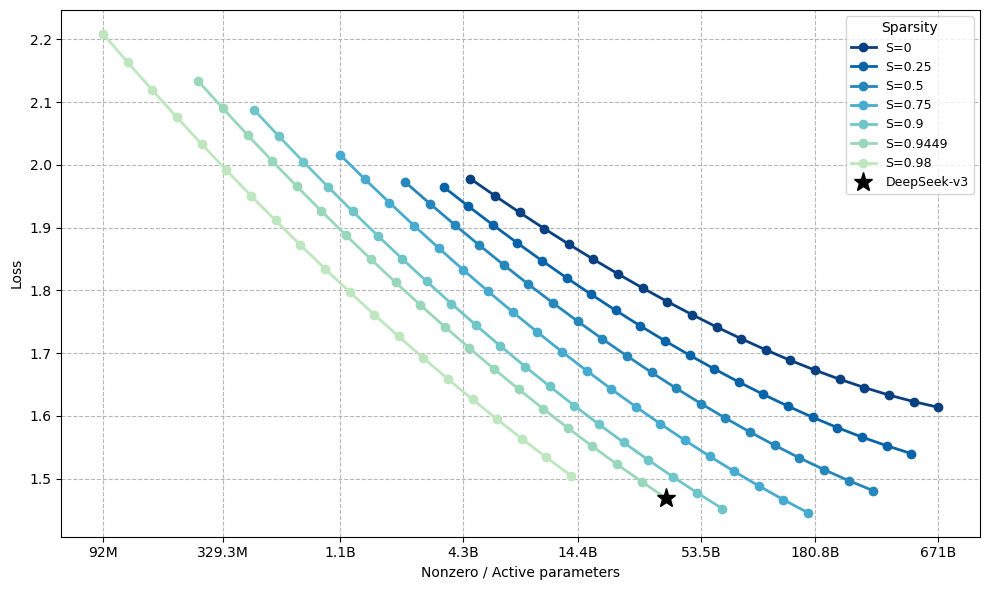}
  \end{subfigure}
  \caption{IsoFLOP plot of Abnar scaling law (left) and our proposed scaling law (right) for a compute budget $3.29e24$ FLOPs of DeepSeek-V3 training.}
  \label{fig:deepseek_isoflop_comparison}
\end{figure*} 

\section{Performance Results and Discussion}
\subsection{Performance Results}\label{performance_result}
In this section we  compare the performance of our proposed scaling law and the previous architecture specific scaling laws for dense and sparse models. We have collected the experimental datasets (number of parameters, number of tokens, and sparsity) used in  \cite{hoffmann2022training, frantar2023scaling, abnar2025parameters} shown in Table \ref{tab:comparison_dataset}.
Hoffman et al.~\cite{hoffmann2022training} provided a list of nine models with number of parameters and training tokens used in their experiments. Frantar et al. ~\cite{frantar2023scaling} provided  four different numbers of nonzero parameters, three different numbers of training tokens, and four different sparsity levels with a total of $48$ experimental runs to collect their experimental results. Abnar et al. ~\cite{abnar2025parameters} used  five different numbers of compute budgets, seven different sparsity levels, and a range of model sizes.  Then we use these experimental datasets to generate plots for their respective scaling laws and use the same datasets to generate plots using our proposed scaling law in Equation \ref{ax} in order to compare them.

\begin{table}[ht]
    \centering
    \resizebox{0.5\textwidth}{!}{
    \begin{tabular}{c c c c c}
    \specialrule{.1em}{0em}{0em}
    Scaling Law & Parameters & Tokens & Sparsities(\%) & Test Models\\
    \specialrule{.1em}{0em}{0em}
    Hoffman \cite{hoffmann2022training} & $400M{-}10T$ & $8B{-}216.2B$ & $-$ & $9$\\
    Frantar \cite{frantar2023scaling} & $1.3M{-}85M$ & $16B{-}65B$ & $0,50,75,87.5$ & $48$\\
    Abnar \cite{abnar2025parameters} & $329M{-}21.2B$ & $15B{-}128B$ & $0,25,50,75,90,95,98$ & $35$\\
    \specialrule{.1em}{0em}{0em}
    \end{tabular}
    }
    \caption{Parameters, tokens, sparsity, and number of test models  collected from the corresponding works.}
    \label{tab:comparison_dataset}
\end{table}

Along with drawing the plots, we measure the prediction Mean Square Error (MSE) between our proposed scaling law and the respective scaling law for better visualization of prediction of our proposed scaling law. 

\begin{figure*}[ht]
  \centering

  \begin{subfigure}[t]{0.33\textwidth}
    \centering
    \includegraphics[width=\linewidth]{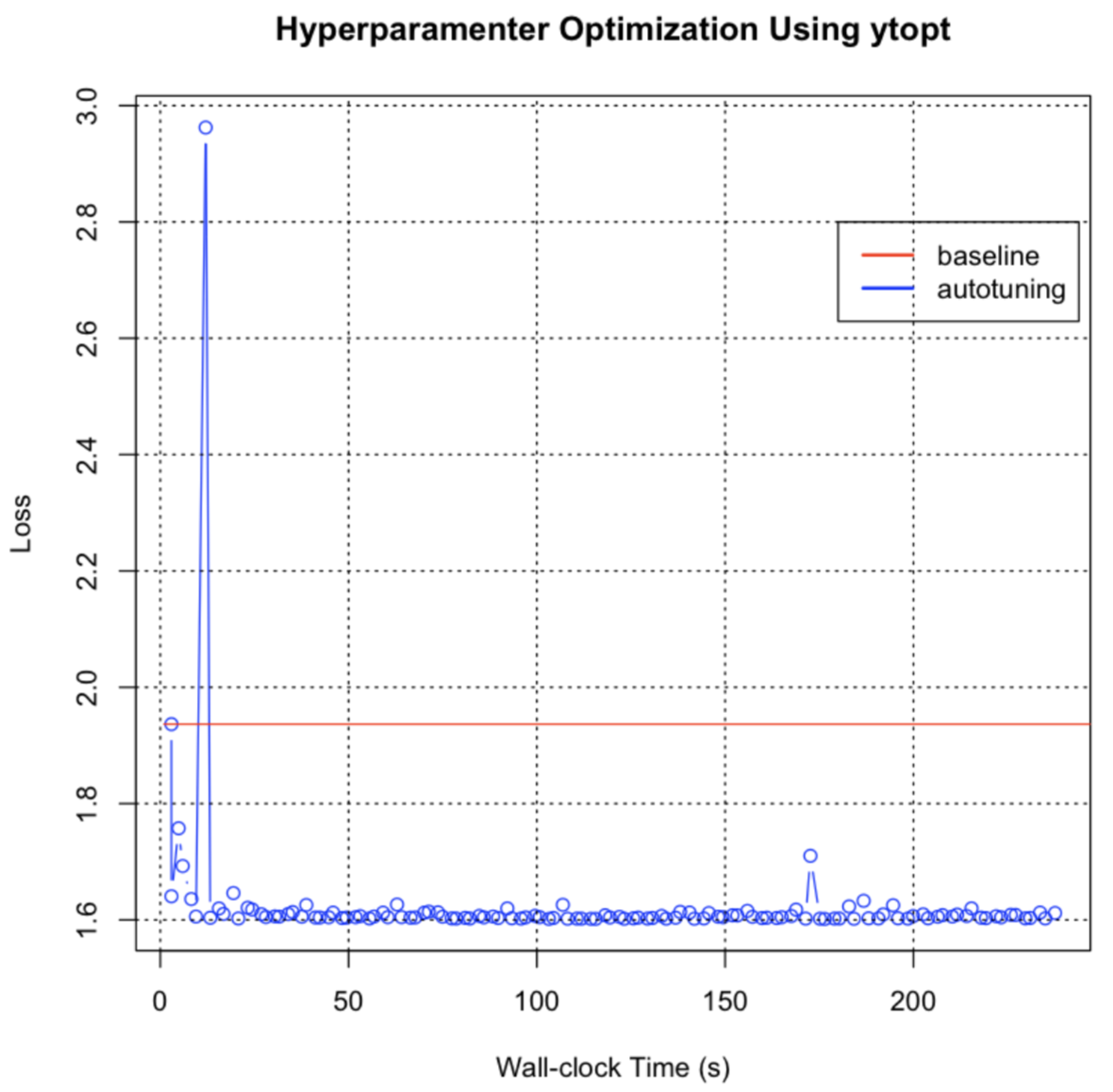}
    \subcaption{Tuning Loss for Equation \ref{hoffman}}
  \end{subfigure}\hfill
  \begin{subfigure}[t]{0.33\textwidth}
    \centering
    \includegraphics[width=\linewidth]{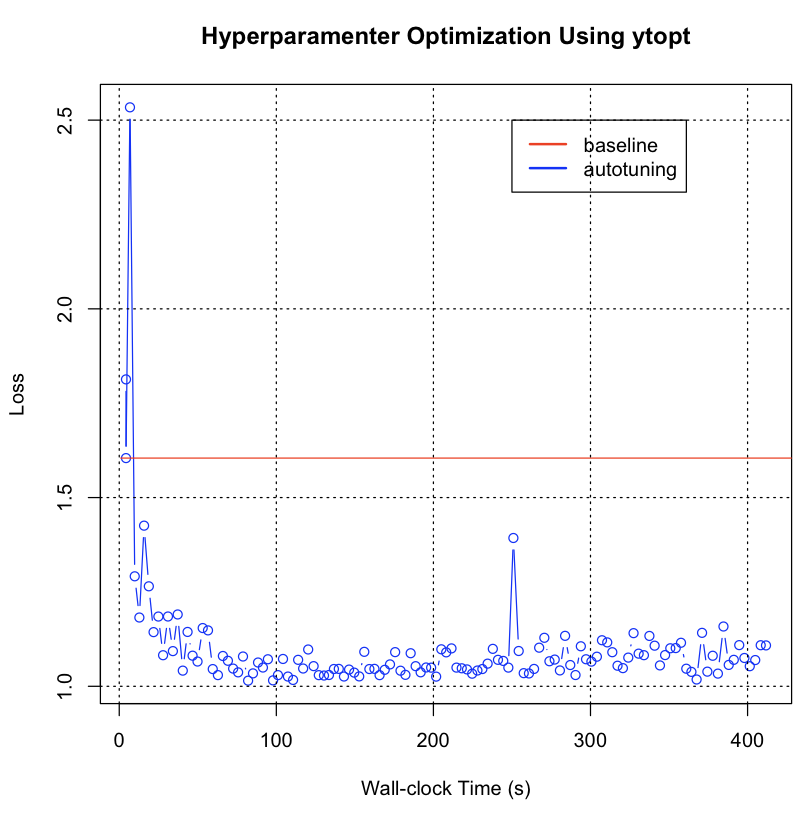}
    \subcaption{Tuning Loss for Equation \ref{frantar}}
  \end{subfigure}
    \begin{subfigure}[t]{0.33\textwidth}
    \centering
    \includegraphics[width=\linewidth]{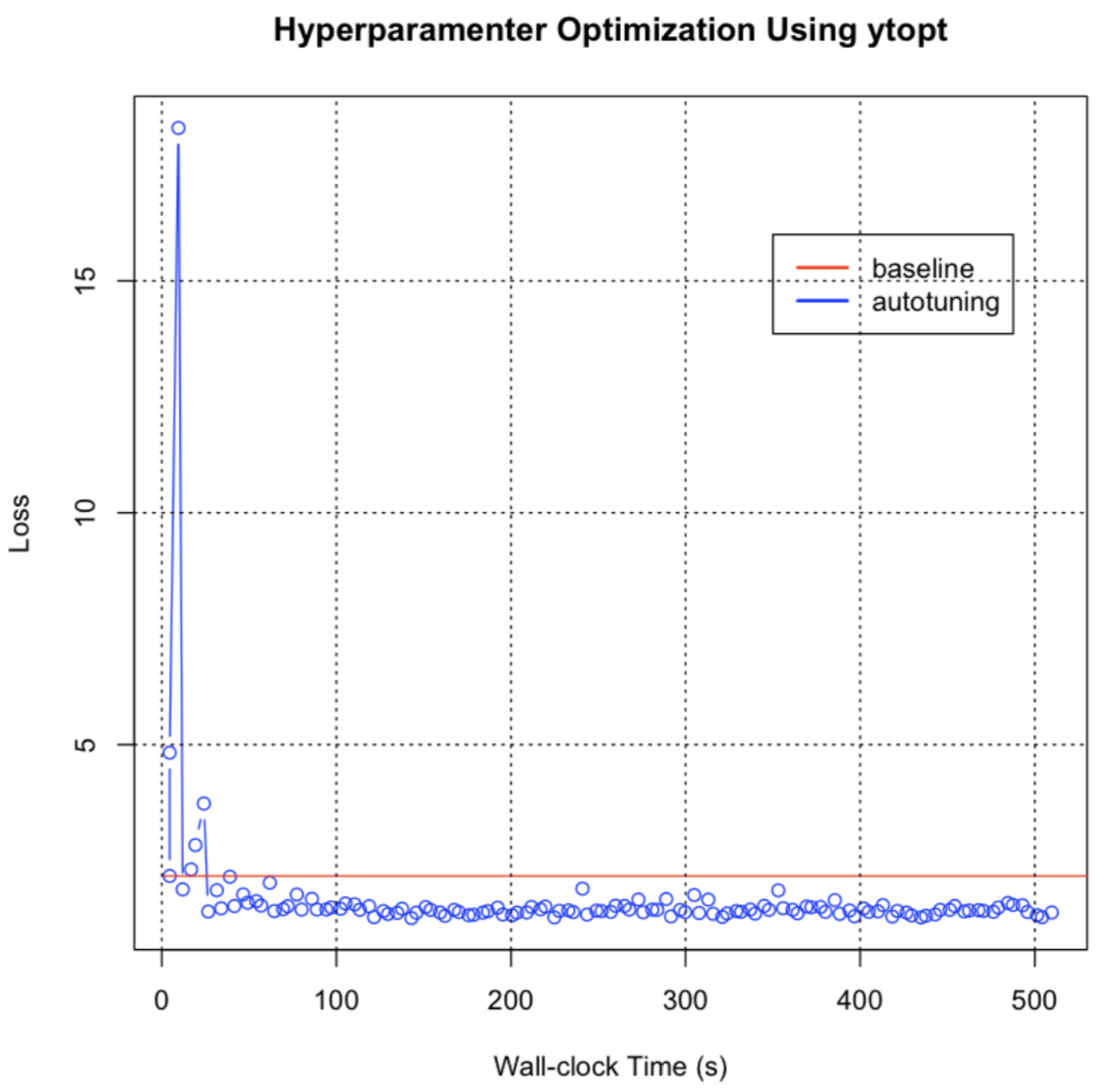}
    \subcaption{Tuning Loss for Equation \ref{abnar}}
  \end{subfigure}

  \caption{Loss calculated by the scaling law coefficients generated by ytopt and grid search. The red lines are the loss using the best scaling law coefficients provided in the respective work, and the blue dots are the loss tuned for different evaluations over time using ytopt.}
  \label{fig:ytopt_vs_grid}
\end{figure*}

First, to compare our proposed scaling law in Equation \ref{ax} with Hoffman scaling law in Equation \ref{hoffman} \cite{hoffmann2022training}, we refer back to the Lemma in section \ref{generalization} where we have shown that if the sparsity $S=0$, Equation \ref{ax} converts into Equation \ref{hoffman}. Table \ref{tab:ax_hoff} shows the fitted coefficient values in Equation \ref{ax} for the dense models from \cite{hoffmann2022training}. Figure \ref{fig:hoffman_vs_proposed_comparison} shows the loss vs compute budget using the dataset in \cite{hoffmann2022training}. On the $x$-axis is compute ($C=6ND$). which depends on model size $N$ and tokens $D$ and hence captures the impact of both. On the  $y$-axis is the calculated loss using both scaling laws. We see an identical plot, indicating that our proposed scaling law behaves the same as the original scaling law for dense models  capturing the dense scaling behavior where increasing compute results in decreasing pretraining loss.

\begin{table} [h]
    \centering
    \begin{tabular}{c c c c c c c c}
    \specialrule{.1em}{0em}{0em}
    Coefficients & $e$ & $a$ & $b$ & $c$ & $\alpha$ & $\beta$ & $\gamma$\\
    \hline
    Values & $1.69$ & $406.4$ & $410.7$ & $0$ & $0.34$ & $0.28$ & $0.01$\\
    \specialrule{.1em}{0em}{0em}
    \end{tabular}
    \caption{Fitted scaling law coefficient values of our proposed scaling law for dense models.} 
    \label{tab:ax_hoff}
\end{table}

To compare our proposed scaling law with Frantar scaling law, we use the data (sparsity, training steps, number of non-zero parameters and training tokens) collected from the 48 models that were used to conduct the experiments in \cite{frantar2023scaling}. Figure \ref{fig:frantar_vs_proposed_comparison} shows that our proposed scaling law captures the almost similar scaling behavior as Frantar scaling law for sparse models as sparsity and number of of nonzero parameters keep increasing.  
The loss prediction is very similar with average $MSE=0.00078$. Table \ref{tab:ax_frantar} shows the fitted parameter values for our proposed scaling law for models from Frantar scaling law experiments \cite{frantar2023scaling}.

\begin{table} [h]
    \centering
    \begin{tabular}{c c c c c c c c}
    \specialrule{.1em}{0em}{0em}
    Coefficients & $e$ & $a$ & $b$ & $c$ & $\alpha$ & $\beta$ & $\gamma$\\
    \hline
    Values & $0.15$ & $86.03$ & $7.90$ & $29.26$ & $0.28$ & $0.08$ & $2.00$\\
    \specialrule{.1em}{0em}{0em}
    \end{tabular}
    \caption{Fitted parameter values of our proposed scaling law for Frantar-style sparse models.} 
    \label{tab:ax_frantar}
\end{table}


We also compared our proposed scaling law with the Abnar scaling law in Equation \ref{abnar} \cite{abnar2025parameters}. 
For this comparison we use the interpolated values of parameters, tokens, and the exact compute budget of the 35 models used to conduct the experiments in \cite{abnar2025parameters}. We have used interpolated values to generate the full dataset because even though the authors have shared a list of compute budget and sparsity levels, no such list was shared for model size and training tokens. So we used the plots shared in \cite{abnar2025parameters} and generated the entire dataset. In Figure \ref{fig:ax_abnar_comparsion} we have drawn compute budget (FLOPs) vs. loss plots to compare the behavior of both scaling laws. We observe that with the increase of the sparsity and compute, the loss decreases, and both scaling laws show a similar pattern. The average prediction MSE between Abnar scaling law and our proposed scaling law is $0.0005$ which is not significant. We present our fitted scaling law coefficient values in Table \ref{tab:ax_abnar}. 

\begin{table} [h]
    \centering
    \begin{tabular}{c c c c c c c c}
    \specialrule{.1em}{0em}{0em}
    Coefficients & $e$ & $a$ & $b$ & $c$ & $\alpha$ & $\beta$ & $\gamma$\\
    \hline
    Values & $0.57$ & $8.26$ & $6324.82$ & $3.57$ & $0.08$ & $0.40$ & $1.19$\\
    \specialrule{.1em}{0em}{0em}
    \end{tabular}
    \caption{Fitted parameter values of our proposed scaling law for MoE-based models.} 
    \label{tab:ax_abnar}
\end{table}



Abnar et al. \cite{abnar2025parameters} also conducted IsoFLOP experiments with the given compute budget of $1e20$ FLOPs for the MoE-based models. In Figure \ref{fig:isoflop_comparison} we have shown the IsoFLOP loss prediction comparison between Abnar scaling law and our proposed scaling law, where each line connects models with different sparsity and model sizes trained under the same amount of compute budget $1e20$ with active parameters as x-axis and training loss as y-axis. In \cite{abnar2025parameters} the authors intentionally cut the plot for active parameters at $4B$ parameters even though the original experiments had active parameters around $26B$. For comparison purpose, we mark that $4B$ parameter spot and generate the entire plot in Figure \ref{fig:isoflop_comparison}. For IsoFLOP experiments \cite{abnar2025parameters}, the authors showed that for fixed sparsity, as we increse number of active parameters, the pretraining loss keeps decreasing until it reaches a sweet spot. Beyond that sweet spot, loss starts increasing again. As seen from Figure \ref{fig:isoflop_comparison}, our proposed scaling law can effectively capture this behavior.     

To demonstrate the effectiveness of our scaling law for larger models, we use DeepSeek-V3 \cite{deepseekai2025deepseekv3technicalreport} which is a MoE-based model with 671B
parameters and 37B of which are activated for each token as an example. The model was trained with $14.8T$ tokens. Using the compute formula $C=6ND$ where $N$ is the number of active parameters and $D$ is the number of training tokens, we measure the total compute, $3.29e24$ used to train DeepSeek-V3. We generate an interpolated IsoFLOP dataset using this compute budget where we restricted the number of total parameters to $671B$ parameters. Figure \ref{fig:deepseek_isoflop_comparison} shows that our proposed scaling law exhibits a very similar pattern to Abnar scaling law for very large models as well.

\subsection{Hyperparameter Optimization for Better Scaling Law Coefficients}
Most of the previous scaling laws discussed in this paper used grid search to find the best values for the scaling law coefficients. While grid search is a commonly used approach, it is still inefficient. As an alternative, we have used ytopt \cite{WU25, wu2025integrating, Ytopt} for scaling law coefficient optimization. ytopt is a Bayesian optimization--based autotuner that builds a surrogate model to explore the promising regions to find the optimal values with fewer evaluations. In our previous work \cite{WU23}, ytopt outperformed grid search, random search, a genetic algorithm, and XGBoost in accuracy and  tuning time. Unlike traditional autotuners, ytopt leverages libEnsemble's \cite{hudson2021libensemble} asynchronous manager-worker framework to run multiple evaluations in parallel, reducing the one-by-one autotuning time. 

To use ytopt for the optimization process for finding the best scaling law coefficients for the previous scaling laws, we use the best scaling law coefficients for the scaling laws in Tables \ref{tab:hoffman_tab}, \ref{tab:frantar_tab_mod}, and \ref{tab:abnar_tab} as the baseline and the default values to define the proper search space for each coefficient. We  carefully define the range for each coefficient for  Equations \ref{hoffman}, \ref{frantar}, and \ref{abnar} and then use ytopt to run the optimization process. The generated hyperparameter optimization results in Figure \ref{fig:ytopt_vs_grid}(a), (b), and (c) illustrate that ytopt achieves better loss predictions than grid search. Since ytopt is a Bayesian optimization-based approach, it is much faster than a grid search-based approach and is overall a better alternative.

\subsection{Discussion}
In summary, we observe that our proposed generalized scaling law Equation \ref{ax} effectively captures the scaling behavior of Equations \ref{hoffman}, \ref{frantar}, and \ref{abnar} for dense and sparse LLMs. The effectiveness is further justified by the small MSE between the existing scaling law and out proposed scaling law predictions. In Figure \ref{fig:isoflop_comparison} and \ref{fig:deepseek_isoflop_comparison} we have seen that our proposed scaling law can capture the IsoFLOP behavior of MoE models as well. So we can conclude that for a given compute budget, our proposed scaling law can be used to estimate the best model hyperparameters (Model size, Tokens and Compute) for a given sparsity or to identify the optimal sparsity for the given model hyperparameters. 

 In this work we have generalized the scaling behaviors of dense and sparse models. Our generalized scaling law extends and covers several existing scaling laws. Its limitations are related to the sparsity factor $c$ and the coefficient $\gamma$, which need to be further optimized for larger LLM models. 
\section{Conclusions}
 As LLMs get larger and more expensive in training, it is crucial for researchers and developers to estimate the expected performance beforehand. Scaling laws thus are critical. While architecture-specific scaling laws give a fine-grained estimation, as more and more new architecture and training methods are being developed, using different scaling laws for each variant becomes impractical. 
Our generalized scaling law covers densely activated models, pruned models, and MoE models. Empirical results show that our generalized approach can effectively capture the scaling behavior of the architectures.  


Because LLMs inference can be memory-bound and compute-intensive, future work will extend the generalized scaling law in training to not only incorporate other scaling laws \cite{kumar2024scaling,busbridge2025distillationscalinglaws,chen2025scaling, wang2024q, thangarasa2023spdf} but also take into account inference \cite{sardana2025chinchillaoptimalaccountinginferencelanguage}, especially recent agentic AI workloads for answering complex questions through reasoning using chain-of-thought prompting. Since the behavior of scaling laws varies based on training and inference approaches, these approaches can be incorporated in the generalized scaling law as well. Unifying different scaling laws into one single representation lays the foundation of architecture-agnostic representation of the scaling behavior of LLMs. 

\section*{Acknowledgments}
This work was supported by DOE ASCR SciDAC RAPIDS and OASIS. We acknowledge the Argonne Leadership Computing Facility for use of Polaris under the project EE-ECP. This material is based upon work supported by the U.S. Department of Energy, Office of Science, under contract number DE-AC02-06CH11357. 

\bibliographystyle{IEEEtran}
\bibliography{ref}


\end{document}